\pdfoutput=1
\documentclass{article} %
\usepackage[preprint]{neurips_2026}

\usepackage{amsmath,amsfonts,bm}

\def\eqref#1{equation~\ref{#1}}

\def\1{\bm{1}}

\DeclareMathAlphabet{\mathsfit}{\encodingdefault}{\sfdefault}{m}{sl}
\SetMathAlphabet{\mathsfit}{bold}{\encodingdefault}{\sfdefault}{bx}{n}

\usepackage{amssymb}
\usepackage{graphicx}
\usepackage[table]{xcolor}
\usepackage{makecell}
\usepackage{array}
\usepackage{siunitx}
\newcolumntype{A}{S[table-format=2.2]}
\newcolumntype{T}{S[table-format=4.1]}

\definecolor{bandtint}{HTML}{EFE4D4}   %
\definecolor{besttint}{HTML}{FCE4DB}   %
\usepackage{booktabs}
\usepackage{multirow}
\usepackage{placeins}
\usepackage{caption}

\usepackage{hyperref}
\usepackage{url}

\title{BASE: Batch-Aware Selection of Experts Using Predicted Removal Error
for Efficient MoE Decoding}

\author{%
  Ali Abbasi \quad Justin Shi \quad Soheil Kolouri \\
  College of Connected Computing, Vanderbilt University \\
  \texttt{\{ali.abbasi, justin.shi, soheil.kolouri\}@vanderbilt.edu}
}

\begin{document}

\maketitle

\begin{abstract}
Large language models continue to grow in parameter count and capacity, making
them increasingly expensive to serve. In large-scale serving systems,
autoregressive decoding is often bottlenecked by transferring model weights
from accelerator high-bandwidth memory into on-chip SRAM. This memory
bottleneck becomes more severe as models and context lengths grow, placing
increasing pressure on memory bandwidth. Mixture-of-experts (MoE) models
reduce computation by activating only a small subset of experts per token, but
this sparsity does not translate directly to batched decoding. Different
requests select different experts; therefore, the combined active set across
many concurrent requests can span a substantial fraction of the expert pool
and require significantly more expert weights to be transferred. Most
expert-reduction techniques make retention decisions independently for each
token and therefore do not address this batch-level expansion. More recently,
batch-aware methods have attempted to coordinate expert use across concurrent
requests and reuse experts already fetched for the batch. Yet their selection
criteria are based primarily on router rankings or expert statistics collected
during calibration. Consequently, these criteria are not directly tied to the
output error caused by dropping an expert, nor do they capture how an expert's
contribution changes across tokens at inference time. We instead rank experts
according to how much their removal would change the MoE-layer output. To
apply this criterion during serving, we train a lightweight linear predictor
during calibration that estimates the expert removal cost for each incoming
token. We also develop custom GPU kernels to efficiently perform cost
prediction, batch-level score aggregation, and expert selection. Across three
MoE architectures, BASE improves the quality--efficiency tradeoff without
retraining. On Qwen3-30B-A3B, it improves average accuracy by 29.5 points over
the strongest baseline at comparable throughput under a tight expert budget.
At a higher expert budget, it is 60\% faster than dense inference while
remaining within 0.4 accuracy points.
\end{abstract}

\section{Introduction}
\label{sec:introduction}
Scaling model size and training compute has consistently improved
language-model performance, but it has also made inference increasingly
expensive~\citep{kaplan_scaling_2020,hoffmann_training_2022}. In large-scale
serving, autoregressive decoding is often constrained by memory bandwidth, as
model weights and a growing key-value (KV) cache must be accessed for every
generated token~\citep{kwon2023efficient,rajbhandari2022deepspeedMoE}. Serving
systems therefore treat decoding as a distinct phase with its own cost
profile~\citep{zhong2024distserve,agrawal2024taming}.
Mixture-of-experts (MoE) architectures reduce per-token computation by
activating only a sparse subset of model parameters
~\citep{jacobs_adaptive_1991,shazeer2017outrageouslylargeneuralnetworks,fedus2021switch,jiang2024mixtral}, yet this sparsity does not
necessarily translate into efficient batched decoding.

This limitation arises because MoE sparsity is defined per token, whereas
serving cost is incurred across the batch. At each decoding step, every active
request contributes one token; although each token selects only a few experts,
their selections can differ substantially, causing the batch-wise union to
grow rapidly and cover a large fraction of the expert
pool~\mbox{\citep{wu2026sere,oncescu2025opportunistic}}. In the memory-bound
decoding
regime, each distinct expert in this union requires its weights to be fetched
from high-bandwidth memory, so latency is governed primarily by the size of the
batch-level active set rather than the sparsity of any individual
token~\citep{rajbhandari2022deepspeedMoE,oncescu2025opportunistic}. Efficient
batched decoding therefore requires controlling which experts are loaded for
the batch as a whole.

Existing methods reduce MoE serving costs by pruning or merging experts before
inference~\citep{lasby2025reap,chen2024retraining,mcmoe2024}, reducing expert
execution per token~\citep{huang2024harder,notall2024experts}, or coordinating
expert use across a
batch~\citep{gupta2024lynx,wu2026sere,oncescu2025opportunistic}. We focus on the
last setting, where the objective is to limit the number of distinct expert
weights fetched at each decode step.

Existing batch methods still leave open how experts should be selected according
to their contribution to the current batch. SERE and OEA construct the active
set from token routing decisions, so an expert selected by only one token can
add an expert to the batch active set even if its contribution to the batch is
small~\citep{wu2026sere,oncescu2025opportunistic}. ExFold instead imposes a
fixed batch budget and combines current router weights with expert output norms
measured during calibration~\citep{wu2026exfold}. However, a fixed calibrated
norm cannot capture how an expert's output varies across tokens in the current
batch. This leads to the central question we study: \textbf{Given a fixed batch
budget, which experts should be fetched to minimize the change in the MoE layer
output?}

To address this question, we introduce BASE, which scores experts using the
squared change in the MoE layer output caused by removing an expert's weighted
contribution. The exact error contains the individual removal costs of omitted
experts and cross terms between experts omitted together. Our analysis shows
that selection is robust to discarding the cross terms, allowing BASE to rank
experts independently and retain those with the largest estimated removal costs
under a fixed batch budget.

The remaining challenge is that removal cost depends on an expert's output for
the current token, which is unavailable before the expert is fetched and
executed. BASE resolves this with a lightweight linear predictor for each
expert, trained during calibration, that estimates output energy from the
current token representation. BASE combines these estimates with router weights,
aggregates removal costs across the batch, and loads the experts with the
largest predicted costs under the fixed budget.

Our contributions are threefold. \textbf{1)} We formulate batch expert selection
as minimizing the change in the MoE layer output under a fixed expert budget,
and show that discarding cross terms has little effect on the selected set.
\textbf{2)} We predict each routed expert's output energy from the current token
representation before execution, combine it with router weights to estimate
removal cost, and use backfill when a routed expert is not loaded. \textbf{3)}
We implement BASE with custom GPU kernels and evaluate it across three MoE
architectures and eight benchmarks. BASE improves average accuracy by 3.0 to
29.5 points over existing methods at comparable speed under tight budgets, while
relaxed settings stay within 0.4 points of dense model quality with 22 to 60\%
higher throughput.

\section{Related Work}
\label{sec:related_work}

Prior work reduces MoE inference cost through offline expert
compression~\citep{lasby2025reap,mcmoe2024,chen2024retraining}, adaptive expert
execution for individual tokens~\citep{huang2024harder,notall2024experts}, or
coordination of expert use across a
batch~\citep{wu2026sere,oncescu2025opportunistic,gupta2024lynx,chen2026dynamicexpertsharingdecoupling,wu2026exfold}.
BASE belongs to the last category, but differs in estimating expert importance
from the predicted output contribution of the current tokens rather than from
router signals or fixed calibration statistics.

\textbf{Offline Expert Pruning and Merging.} Offline methods permanently reduce
the global expert pool through pruning, merging, decomposition, or
regularization~\citep{yang2024moe,liu2024efficient,mcmoe2024,chen2024retraining,muzio2024seer,jaiswal_finding_2025}.
REAP is particularly relevant because it combines router weights with expert
output magnitudes and relates this score to reconstruction
error~\citep{lasby2025reap}. Its pruning decisions, however, are fixed from
calibration data. BASE instead estimates removal costs for the current tokens
and chooses the active expert set separately for each decode batch.

\textbf{Token-Level Adaptive Expert Execution.} Other methods adapt the number
of experts executed for individual tokens using router confidence, sensitivity,
or learned importance
criteria~\citep{huang2024harder,li2023adaptive,zhong2024adapmoe,notall2024experts,huang2025modes,yang2025faster}.
Because different tokens can retain different experts, reducing computation per
token does not directly control the number of distinct experts fetched for the
batch.

\textbf{Batch-Aware Expert Selection.} SERE and OEA construct the batch active
set from the union of preferred experts across
tokens~\citep{wu2026sere,oncescu2025opportunistic}. OEA fills missing routes
with experts already loaded for the batch, while SERE uses calibrated expert
similarities to choose replacements. Because their active set is formed from a
union, its size depends on routing overlap and can grow when only a few tokens
request additional experts. Lynx instead uses router confidence and expert
popularity within the current batch to reduce the active
set~\citep{gupta2024lynx}, but does not estimate the output change caused by
removing an expert. ExFold is closest to BASE because it combines a fixed batch
budget with output information~\citep{wu2026exfold}. However, it uses expert
output norms measured during calibration, whereas BASE predicts output energy
from the current token representation. ExFold also transfers the contribution of
a removed expert to a retained expert using a calibrated scalar, while BASE uses
backfill among the token's preferred experts.

\textbf{Expert Offloading and Prefetching.} Systems such as Fiddler and
Pre-gated MoE reduce serving cost through CPU and GPU placement, offloading, or
prefetching~\citep{kamahori2025fiddlercpugpuorchestrationfast,hwang2024pregatedmoealgorithmsystemcodesign,balmau2025accelerating}.
BASE instead reduces the number of expert weights accessed from GPU
high-bandwidth memory during decoding.

\section{Method}
\label{sec:method}

\subsection{Problem Formulation}
\label{sec:problem_formulation}

Consider an MoE layer with $N$ routed experts
$E_1,\ldots,E_N:\mathbb{R}^{d}\rightarrow\mathbb{R}^{d}$.
For token $t$ with input $z_t$, the router selects $K$ experts
$R_t\subseteq[N]$ with weights $w_{t,u}$, producing
$y_t=\sum_{u\in R_t}w_{t,u}E_u(z_t)$. We assume token-choice routing, omit shared experts from the notation, and let
$\ell$ index the layer~\citep{zhou2022mixture,dai2024deepseekmoeultimateexpertspecialization}.

\textbf{Batched decoding.} At each decode step, the batch $\mathcal{B}$
contains one token from each active sequence. Although each token routes to
only $K$ experts, the batch can require
$\left|\bigcup_{t\in\mathcal{B}}R_t\right|$ distinct experts, which can be
much larger than $K$.

MoE decoding is typically
memory-bound~\citep{rajbhandari2022deepspeedMoE,oncescu2025opportunistic}. We
model the layer latency as
\begin{equation}
    T_\ell
    \approx
    \tau_{\mathrm{fetch}}|A_\ell|
    +
    \tau_{\mathrm{comp}}
    \sum_{t\in\mathcal{B}}|R_t'|,
    \label{eq:moe_latency}
\end{equation}
where $A_\ell$ is the set of experts fetched for the batch and $R_t'$ is the
set executed for token $t$. Here, $\tau_{\mathrm{fetch}}$ is the per-expert
weight-transfer cost and $\tau_{\mathrm{comp}}$ is the per-token, per-expert
computation cost. In the memory-bound regime, the first term
dominates, making latency primarily dependent on
$|A_\ell|$~\citep{oncescu2025opportunistic}. On our inference stack, decode
time is approximately linear in the expert budget over $4\leq M\leq32$, with
$R^2\geq0.996$ across all three architectures
(Appendix~\ref{app:latency_budget}). We therefore constrain the active set to
$|A_\ell|\le M$ and ask which experts should be fetched for the current batch.

Unlike standard token
routing~\citep{shazeer2017outrageouslylargeneuralnetworks,fedus2021switch} and
union-based methods such as SERE and
OEA~\citep{wu2026sere,oncescu2025opportunistic}, we directly constrain the
number of experts fetched and select the active set according to the resulting
change in the MoE output.

\begin{figure}[t]
\centering
\includegraphics[width=\linewidth]{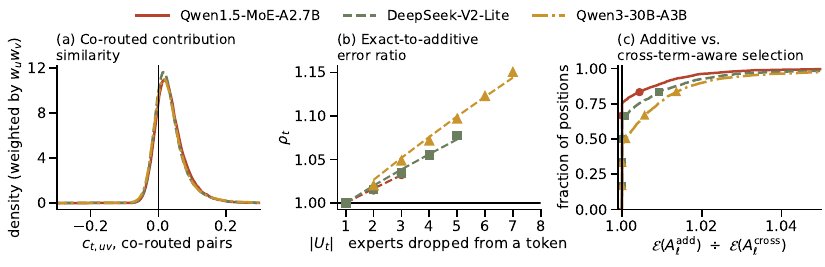}
\caption{\textbf{Cross terms affect error magnitude but have limited impact on
expert selection.} \textbf{(a)} $w_{t,u}w_{t,v}$-weighted distributions of cosine similarity
between expert contributions routed to the same token. \textbf{(b)} Exact-to-additive
error ratio as the literal number of omitted experts increases. The active set
is the batch-wise union of each token's top-1 expert; dashed curves show
one-parameter fits. \textbf{(c)} Exact-error ratio between additive and cross-term-aware
selection under the same fixed budget $M=12$. Panels (a) and (b) exclude the
first and last MoE layers.}
\label{fig:expert_geometry}
\end{figure}

\subsection{What Is the Cost of Leaving an Expert Out?}
\label{sec:expert_omission_cost}

For token $t$, define the weighted contribution
$a_{t,u}\triangleq w_{t,u}E_u(z_t)$, and let
$U_t\triangleq R_t\setminus A_\ell$ denote its omitted experts. The resulting
batch output error is
\begin{equation}
    \mathcal{E}(A_\ell)
    =
    \sum_{t\in\mathcal{B}}
    \left\|
    \sum_{u\in U_t}a_{t,u}
    \right\|_2^2
    =
    \underbrace{\sum_{t\in\mathcal{B}}\sum_{u\in U_t}\|a_{t,u}\|_2^2}_{\text{individual damage}}
    +
    \underbrace{2\sum_{t\in\mathcal{B}}\sum_{u<v,\,u,v\in U_t}\langle a_{t,u},a_{t,v}\rangle}_{\text{cross terms}}.
    \label{eq:error_decomposition}
\end{equation}

The individual removal cost of expert $u$, aggregated across the batch, is
\begin{equation}
    D_u
    \triangleq
    \sum_{t\in\mathcal{B},\,u\in R_t}
    w_{t,u}^2\|E_u(z_t)\|_2^2.
    \label{eq:individual_expert_damage}
\end{equation}
Ignoring the cross terms gives
$\mathcal{E}_{\mathrm{add}}(A_\ell)=\sum_{u\notin A_\ell}D_u$, which under
$|A_\ell|=M$ is minimized by retaining the $M$ experts with the largest $D_u$.
Thus, additive selection reduces the problem to ranking experts by their
individual removal costs.

However, the additive objective is not generally identical to the exact
objective. When several experts are omitted for the same token, their
contributions interact through cross terms of the form
$\langle a_{t,u},a_{t,v}\rangle
=\|a_{t,u}\|_2\|a_{t,v}\|_2\cos\theta_{t,uv}$. Large or systematically
structured cross terms could invalidate an independent per-expert score. In
that case, the cost of omitting expert $u$ would depend strongly on which
other experts are omitted alongside it. This leads to the first question our method must answer: \textbf{Can we safely
ignore the cross terms when selecting which experts to retain?}

A second question follows from the same geometry. SERE compensates for an unavailable expert by replacing it with a similar expert already loaded for the batch~\citep{wu2026sere}. Such substitution is useful only when the corresponding expert outputs point in sufficiently similar directions. This leads to our second question: \textbf{Can the output of one active expert serve as a reliable surrogate for an unavailable expert?}

Therefore, before designing either the selection rule or the treatment of
omitted experts, we measure the geometry of expert outputs produced for the
same token. Figure~\ref{fig:expert_geometry} summarizes this analysis.

\begin{figure}[t]
\centering
\includegraphics[width=\linewidth]{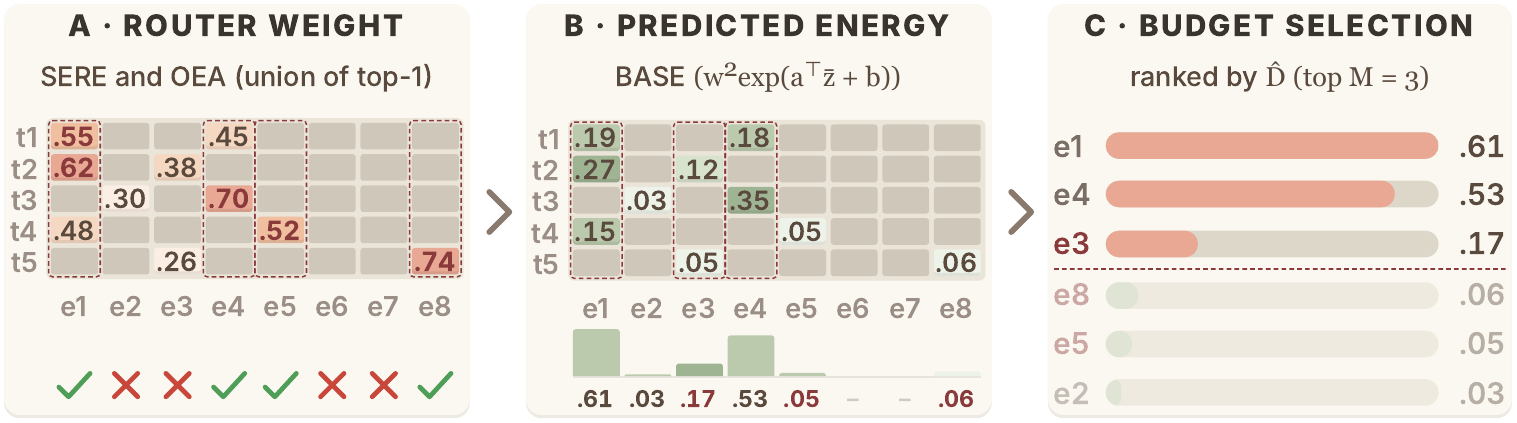}
\caption{\textbf{Binary union and budgeted selection produce different active
sets.} Constructed example with five tokens, eight experts, and two routed
experts per token. \textbf{(a)} SERE and OEA fetch the union of each token's
top-1 expert, shown by dashed columns, producing four active experts.
\textbf{(b)} BASE computes each routed pair's predicted removal cost,
$w_{t,u}^2\widehat e_u(z_t)$, and sums these values across tokens to obtain the
batch-level expert score $\widehat D_u$. \textbf{(c)} BASE fetches the $M=3$
highest-scoring experts. Experts $e_5$ and $e_8$ enter the binary union despite
low predicted removal costs, while $e_3$ is selected by BASE despite never
being a token's top-1 expert.}
\label{fig:teaser}
\end{figure}

\subsection{Expert Selection Is Robust to Discarding Cross Terms}
\label{sec:cross_term_robustness}

The exact objective in Equation~\ref{eq:error_decomposition} contains cross
terms between omitted expert contributions. Discarding them is useful only if
it does not substantially change which experts should be retained. We examine
this across all three architectures using the ground-truth outputs of experts
in each token's original top-$K$ routing set. For two experts $u$ and $v$
routed to token $t$, let $c_{t,uv}$ denote the cosine similarity between their
outputs.

\textbf{Co-routed expert outputs are nearly orthogonal.} Figure~\ref{fig:expert_geometry}(a)
shows that these similarities concentrate near zero, with weighted mean
absolute values of $0.045$, $0.040$, and $0.043$ for Qwen1.5-MoE-A2.7B,
DeepSeek-V2-Lite, and Qwen3-30B-A3B, respectively.

Near-orthogonality does not guarantee that several small cross terms cannot
accumulate. We therefore measure their combined effect using the ratio between
exact and additive error:
\begin{equation}
    \rho_t
    \triangleq
    \frac{
        \left\|
        \sum_{u\in U_t}a_{t,u}
        \right\|_2^2
    }{
        \sum_{u\in U_t}\|a_{t,u}\|_2^2
    },
    \label{eq:exact_additive_ratio}
\end{equation}
where $U_t=R_t\setminus A_\ell$ is the set of omitted experts. A value of
$\rho_t=1$ corresponds to exact additivity; values above or below one indicate
amplification or cancellation from the cross terms.

\textbf{Cross-term bias is small and predictable.} For this analysis, we
construct $A_\ell$ as the union of each token's top-1 expert, matching the
active-set construction used by SERE and OEA.
Figure~\ref{fig:expert_geometry}(b) groups tokens by the number of omitted
experts, $|U_t|$, and reports their mean $\rho_t$. The ratio increases
gradually with $|U_t|$, showing that cross terms introduce a modest bias in the
additive error as more experts are omitted. Each point corresponds to tokens
with the same integer value of $|U_t|$. The dashed curves show one-parameter
fits, with fitting details provided in Appendix~\ref{app:crossterm_fit}.

Thus, the additive objective does not perfectly predict the magnitude of the
exact error. The more important question for our method is whether this bias
changes which experts should be retained.

\textbf{Additive scoring preserves selection quality.}
Figure~\ref{fig:expert_geometry}(c) evaluates the fixed-budget selection
problem. At each decode position, we compare the additive selection
$A_\ell^{\mathrm{add}}$ with a selection $A_\ell^{\mathrm{cross}}$ that accounts
for the cross terms, using the same budget $M=12$. We construct
$A_\ell^{\mathrm{cross}}$ greedily by repeatedly adding the expert that
produces the largest reduction in the exact objective. This accounts for all
cross terms, although it does not exhaustively search all subsets of size $M$.

Both selections are evaluated using the exact error from
Equation~\ref{eq:error_decomposition}. Panel (c) reports
$\mathcal{E}(A_\ell^{\mathrm{add}})/\mathcal{E}(A_\ell^{\mathrm{cross}})$,
which measures the additional exact error incurred by additive selection
relative to the cross-term selection. The 95th-percentile ratios are $1.018$,
$1.025$, and $1.032$ for Qwen1.5-MoE-A2.7B, DeepSeek-V2-Lite, and Qwen3-30B-A3B,
respectively. Thus, in 95\% of measured layer and decode position pairs,
discarding the cross terms increases the exact error by no more than $1.8\%$,
$2.5\%$, and $3.2\%$.

These results show an important distinction: \textbf{cross terms affect the
magnitude of the error, but have little effect on which experts should be
retained.}

The same geometry also informs how unavailable experts should be handled.
SERE redirects an unavailable route to a similar expert already in the batch
active set~\citep{wu2026sere}, which assumes that one expert's output can
approximate another's.

We test this assumption directly. Suppose that an omitted expert $u$ is
replaced by an active expert $v$ while retaining $u$'s routing weight. The
substitution error is
\begin{equation}
    \left\|w_{t,u}E_u(z_t)-w_{t,u}E_v(z_t)\right\|_2^2
    =
    w_{t,u}^2\left\|E_u(z_t)-E_v(z_t)\right\|_2^2.
    \label{eq:substitution_error}
\end{equation}
By comparison, dropping expert $u$ produces error
$w_{t,u}^2\|E_u(z_t)\|_2^2$. Substitution is better only if
$2\langle E_u(z_t),E_v(z_t)\rangle>\|E_v(z_t)\|_2^2$. When the two output
norms are comparable, this approximately requires $c_{t,uv}>1/2$, far above
the similarities observed in Figure~\ref{fig:expert_geometry}(a). This supports
using the additive score $D_u$ for selection without treating one expert's
output as a surrogate for another. Table~\ref{tab:fill_rule_ablation} further
shows that substitution performs worse than simply dropping the unavailable
contribution.

\subsection{Predicting Expert Output Energy}
\label{sec:energy_prediction}

Computing $D_u$ requires $\|E_u(z_t)\|_2^2$, which is unavailable before
expert execution. We therefore predict this quantity from the current token
representation. For each expert $u$, calibration examples use the target
$s_u(z)=\log(\|E_u(z)\|_2^2+\epsilon)$. After normalizing the input as
$\bar z=z/\|z\|_2$, we fit the linear model
\begin{equation}
    \widehat s_u(z)=a_u^\top\bar z+b_u,\qquad
    \widehat e_u(z)=\exp\!\left(\widehat s_u(z)\right),\qquad
    \widehat D_u=\sum_{t\in\mathcal{B},\,u\in R_t}w_{t,u}^2\widehat e_u(z_t).
    \label{eq:log_energy_predictor}
\end{equation}
Each predictor is fit independently by ridge regression on routed token-expert
pairs collected during calibration. At inference, BASE selects the $M$ experts
with the largest $\widehat D_u$.

We also evaluate a fixed mean expert energy and a predictor with a shared
nonlinear residual in Section~\ref{sec:selection_score_ablation}; the linear
predictor gives the best accuracy--efficiency tradeoff.

\subsection{Inference Procedure}
\label{sec:inference_procedure}

At each MoE layer, BASE computes $\widehat D_u$ for the routed experts and
loads the $M$ experts with the largest scores (Figure~\ref{fig:teaser}). Each
token then executes its $K$ highest-ranked experts within this active set. If
one of its original routes is unavailable, it backfills the slot with its
next-preferred expert already in the active set, similar to
OEA~\citep{oncescu2025opportunistic}. Backfill therefore requires additional
expert computation but no additional expert weight transfer.

Backfilled experts use their original router scores, with the normalization
denominator computed from the token's original top-$K$ experts rather than
renormalized over the executed experts as in OEA. Selection- and
fill-rule ablations are given in Sections~\ref{sec:selection_score_ablation}
and~\ref{sec:fill_rule_ablation}; kernel details are provided in
Appendix~\ref{app:custom_kernels}.

\section{Experiments}
\label{sec:experiments}

\begin{table}[!t]
\centering
\small
\setlength{\tabcolsep}{3pt}
\resizebox{.96\textwidth}{!}{
\begin{tabular}{c @{\hskip 4pt} c @{\hskip 5pt} l @{\hskip 9pt} AAA @{\hskip 9pt} AAA @{\hskip 9pt} AA @{\hskip 9pt} A @{\hskip 9pt} T}
\toprule[1.6pt]
& & \multirow{2}{*}{\textbf{Methods \textbackslash{} Tasks}}
  & \multicolumn{3}{c}{\textbf{Exam}}
  & \multicolumn{3}{c}{\textbf{Math}}
  & \multicolumn{2}{c}{\textbf{Code}}
  & \multicolumn{1}{c}{\multirow{2}{*}{\makecell[c]{\textbf{Avg.} \\ (Acc. $\uparrow$)}}}
  & \multicolumn{1}{c}{\multirow{2}{*}{\makecell[c]{\textbf{tok/s} \\ ($\uparrow$)}}} \\
\cmidrule(lr){4-6} \cmidrule(lr){7-9} \cmidrule(lr){10-11}
& & & {\textbf{cmmlu}} & {\textbf{boolq}} & {\textbf{bbh}} & {\textbf{math}} & {\textbf{gsm8k}} & {\textbf{math$_{401}$}} & {\textbf{heval}} & {\textbf{mbpp}} & & \\
\midrule[0.75pt]
\multirow{11}{*}{\rotatebox[origin=c]{90}{\small\textbf{Qwen3-30B-A3B}}} & & Dense & 84.79 & 89.82 & 76.93 & 72.34 & 88.40 & 80.55 & 88.41 & 78.80 & 82.50 & 845.3 \\
\cmidrule[0.6pt](lr){2-13}
 & \multirow{5}{*}{\rotatebox[origin=c]{90}{\makecell[c]{\scriptsize\textbf{Restricted}\\[-1pt]\scriptsize$\approx S\!=\!1$}}} & SERE \textcolor{gray}{$_{S=1}$} & 51.04 & 74.07 & 40.20 & \bfseries 29.26 & 38.97 & 40.90 & 16.46 & 9.40 & 37.54 & \bfseries 1452.9 \\
 &  & OEA \textcolor{gray}{$_{k_0=1}$} & 64.64 & \bfseries 86.51 & 43.69 & 27.78 & 33.28 & 34.66 & \bfseries 43.90 & 9.60 & 43.01 & 1372.5 \\
 &  & Lynx \textcolor{gray}{$_{\alpha=0.47,\ \beta=1}$} & \bfseries 68.46 & 84.01 & 51.76 & 19.80 & 38.29 & 65.34 & 29.88 & 17.40 & \bfseries 46.87 & 1304.9 \\
 &  & ExFold \textcolor{gray}{$_{D=8,\ P=8}$} & 40.92 & 67.52 & \bfseries 51.93 & 15.42 & \bfseries 68.16 & \bfseries 70.82 & 15.24 & \bfseries 25.20 & 44.40 & 1395.4 \\
 &  & BASE (ours) \textcolor{gray}{$_{M=10}$} & \cellcolor{besttint}\bfseries 79.46 & \cellcolor{besttint}\bfseries 89.76 & \cellcolor{besttint}\bfseries 73.55 & \cellcolor{besttint}\bfseries 63.94 & \cellcolor{besttint}\bfseries 89.01 & \cellcolor{besttint}\bfseries 80.55 & \cellcolor{besttint}\bfseries 75.00 & \cellcolor{besttint}\bfseries 59.80 & \cellcolor{besttint}\bfseries 76.38 & \cellcolor{besttint}\bfseries 1463.5 \\
\cmidrule[0.6pt](lr){2-13}
 & \multirow{5}{*}{\rotatebox[origin=c]{90}{\makecell[c]{\scriptsize\textbf{Relaxed}\\[-1pt]\scriptsize$\approx S\!=\!2$}}} & SERE \textcolor{gray}{$_{S=2}$} & 78.64 & 89.20 & 64.88 & 64.68 & 79.68 & 76.31 & 78.66 & 29.80 & 70.23 & 1223.5 \\
 &  & OEA \textcolor{gray}{$_{k_0=2}$} & 80.65 & \cellcolor{besttint}\bfseries 90.61 & 65.56 & 63.96 & 77.10 & 71.82 & 82.93 & 28.60 & 70.15 & 1190.7 \\
 &  & Lynx \textcolor{gray}{$_{\alpha=3,\ \beta=2}$} & \cellcolor{besttint}\bfseries 84.62 & 90.31 & \bfseries 75.47 & \bfseries 71.78 & 87.72 & \cellcolor{besttint}\bfseries 83.04 & \cellcolor{besttint}\bfseries 87.20 & \bfseries 73.60 & \bfseries 81.72 & 1047.9 \\
 &  & ExFold \textcolor{gray}{$_{D=18,\ P=8}$} & 83.67 & \bfseries 90.34 & 74.91 & 70.68 & \bfseries 88.86 & \bfseries 82.04 & 77.44 & 70.60 & 79.82 & \bfseries 1225.8 \\\
 &  & BASE (ours) \textcolor{gray}{$_{M=16}$} & \bfseries 83.94 & 90.34 & \cellcolor{besttint}\bfseries 76.85 & \cellcolor{besttint}\bfseries 72.76 & \cellcolor{besttint}\bfseries 90.14 & 81.55 & \bfseries 87.20 & \cellcolor{besttint}\bfseries 77.00 & \cellcolor{besttint}\bfseries 82.47 & \cellcolor{besttint}\bfseries 1350.2 \\
\midrule[0.75pt]
\multirow{11}{*}{\rotatebox[origin=c]{90}{\small\textbf{Qwen1.5-MoE-A2.7B}}} & & Dense & 69.35 & 80.73 & 36.68 & 14.82 & 53.83 & 60.85 & 48.17 & 34.40 & 49.85 & 1315.6 \\
\cmidrule[0.6pt](lr){2-13}
 & \multirow{5}{*}{\rotatebox[origin=c]{90}{\makecell[c]{\scriptsize\textbf{Restricted}\\[-1pt]\scriptsize$\approx S\!=\!1$}}} & SERE \textcolor{gray}{$_{S=1}$} & 46.33 & 78.35 & 32.62 & 5.46 & 24.94 & 46.88 & 6.10 & 13.80 & 31.81 & 1720.1 \\
 &  & OEA \textcolor{gray}{$_{k_0=1}$} & 67.21 & 79.85 & \bfseries 36.99 & \bfseries 9.44 & \bfseries 41.77 & 58.10 & \bfseries 32.93 & 29.40 & \bfseries 44.46 & 1711.8 \\
 &  & Lynx \textcolor{gray}{$_{\alpha=0.7,\ \beta=1}$} & 67.74 & 80.34 & 34.17 & 6.12 & 21.76 & 54.11 & 22.56 & 24.20 & 38.88 & 1677.6 \\
 &  & ExFold \textcolor{gray}{$_{D=11,\ P=4}$} & \cellcolor{besttint}\bfseries 69.39 & \cellcolor{besttint}\bfseries 80.58 & 36.62 & 8.34 & 25.63 & \cellcolor{besttint}\bfseries 61.35 & 25.00 & \bfseries 30.40 & 42.16 & \bfseries 1744.8 \\
 &  & BASE (ours) \textcolor{gray}{$_{M=11}$} & \bfseries 69.28 & \bfseries 80.43 & \cellcolor{besttint}\bfseries 37.60 & \cellcolor{besttint}\bfseries 13.04 & \cellcolor{besttint}\bfseries 47.23 & \bfseries 59.35 & \cellcolor{besttint}\bfseries 39.63 & \cellcolor{besttint}\bfseries 32.80 & \cellcolor{besttint}\bfseries 47.42 & \cellcolor{besttint}\bfseries 1799.6 \\
\cmidrule[0.6pt](lr){2-13}
 & \multirow{5}{*}{\rotatebox[origin=c]{90}{\makecell[c]{\scriptsize\textbf{Relaxed}\\[-1pt]\scriptsize$\approx S\!=\!2$}}} & SERE \textcolor{gray}{$_{S=2}$} & 65.42 & 80.40 & 35.77 & 12.80 & 46.78 & 57.36 & 39.02 & 28.60 & 45.77 & 1476.7 \\
 &  & OEA \textcolor{gray}{$_{k_0=2}$} & 69.16 & 80.00 & 36.23 & \bfseries 14.98 & \bfseries 51.78 & \bfseries 60.10 & 48.78 & 30.80 & \bfseries 48.98 & 1460.9 \\
 &  & Lynx \textcolor{gray}{$_{\alpha=3,\ \beta=2}$} & \cellcolor{besttint}\bfseries 69.44 & \cellcolor{besttint}\bfseries 80.73 & 36.35 & 14.20 & 51.40 & 55.36 & 41.46 & 31.80 & 47.59 & \bfseries 1479.6 \\
 &  & ExFold \textcolor{gray}{$_{D=21,\ P=4}$} & \bfseries 69.35 & \bfseries 80.73 & \cellcolor{besttint}\bfseries 37.22 & 14.26 & 45.56 & \cellcolor{besttint}\bfseries 60.35 & \cellcolor{besttint}\bfseries 51.83 & \bfseries 32.20 & 48.94 & 1468.6 \\
 &  & BASE (ours) \textcolor{gray}{$_{M=18}$} & 69.26 & 80.64 & \bfseries 37.07 & \cellcolor{besttint}\bfseries 15.94 & \cellcolor{besttint}\bfseries 52.69 & 59.85 & \bfseries 50.00 & \cellcolor{besttint}\bfseries 32.60 & \cellcolor{besttint}\bfseries 49.76 & \cellcolor{besttint}\bfseries 1609.8 \\
\midrule[0.75pt]
\multirow{11}{*}{\rotatebox[origin=c]{90}{\small\textbf{DeepSeek-V2-Lite}}} & & Dense & 52.93 & 80.76 & 49.14 & 22.52 & 58.53 & 69.08 & 50.00 & 45.80 & 53.59 & 1056.4 \\
\cmidrule[0.6pt](lr){2-13}
 & \multirow{5}{*}{\rotatebox[origin=c]{90}{\makecell[c]{\scriptsize\textbf{Restricted}\\[-1pt]\scriptsize$\approx S\!=\!1$}}} & SERE \textcolor{gray}{$_{S=1}$} & 24.04 & 67.80 & 33.18 & 5.12 & 26.08 & 44.39 & 6.10 & 3.80 & 26.31 & \bfseries 1633.2 \\
 &  & OEA \textcolor{gray}{$_{k_0=1}$} & 36.41 & \bfseries 79.02 & 43.76 & 14.52 & \bfseries 48.45 & 63.84 & \bfseries 35.37 & \bfseries 33.00 & \bfseries 44.30 & 1589.2 \\
 &  & Lynx \textcolor{gray}{$_{\alpha=0.5,\ \beta=1}$} & 35.66 & 75.44 & 35.66 & 5.42 & 11.30 & 52.62 & 3.66 & 18.00 & 29.72 & 1514.8 \\
 &  & ExFold \textcolor{gray}{$_{D=9,\ P=6}$} & \bfseries 42.17 & 78.26 & \bfseries 44.82 & \bfseries 15.20 & 46.78 & \cellcolor{besttint}\bfseries 69.08 & 19.51 & 30.20 & 43.25 & 1621.7 \\
 &  & BASE (ours) \textcolor{gray}{$_{M=10}$} & \cellcolor{besttint}\bfseries 45.86 & \cellcolor{besttint}\bfseries 79.11 & \cellcolor{besttint}\bfseries 46.64 & \cellcolor{besttint}\bfseries 19.46 & \cellcolor{besttint}\bfseries 56.63 & \bfseries 68.83 & \cellcolor{besttint}\bfseries 46.34 & \cellcolor{besttint}\bfseries 44.40 & \cellcolor{besttint}\bfseries 50.91 & \cellcolor{besttint}\bfseries 1679.4 \\
\cmidrule[0.6pt](lr){2-13}
 & \multirow{5}{*}{\rotatebox[origin=c]{90}{\makecell[c]{\scriptsize\textbf{Relaxed}\\[-1pt]\scriptsize$\approx S\!=\!2$}}} & SERE \textcolor{gray}{$_{S=2}$} & 43.73 & 79.85 & 44.30 & 20.54 & 53.07 & 64.59 & 44.51 & 34.80 & 48.17 & 1365.2 \\
 &  & OEA \textcolor{gray}{$_{k_0=2}$} & 46.81 & 80.43 & \bfseries 48.04 & 21.52 & 56.79 & 68.08 & \cellcolor{besttint}\bfseries 53.05 & 41.80 & 52.06 & 1362.1 \\
 &  & Lynx \textcolor{gray}{$_{\alpha=0.7,\ \beta=1}$} & 47.05 & \bfseries 80.86 & 43.71 & 21.36 & 51.10 & 67.83 & 45.12 & 40.40 & 49.68 & 1302.3 \\
 &  & ExFold \textcolor{gray}{$_{D=19,\ P=6}$} & \bfseries 49.50 & \cellcolor{besttint}\bfseries 81.31 & 47.85 & \bfseries 22.68 & \cellcolor{besttint}\bfseries 60.58 & \bfseries 68.83 & 50.00 & \cellcolor{besttint}\bfseries 46.80 & \bfseries 53.44 & \bfseries 1381.3 \\
 &  & BASE (ours) \textcolor{gray}{$_{M=19}$} & \cellcolor{besttint}\bfseries 51.01 & 80.28 & \cellcolor{besttint}\bfseries 48.48 & \cellcolor{besttint}\bfseries 24.44 & \bfseries 59.82 & \cellcolor{besttint}\bfseries 70.32 & \bfseries 52.44 & \bfseries 45.40 & \cellcolor{besttint}\bfseries 54.02 & \cellcolor{besttint}\bfseries 1435.7 \\
\bottomrule[1.6pt]
\end{tabular}
}
\caption{Main quality and throughput comparison across three MoE architectures under restricted and relaxed operating regimes. Avg. is the average across eight benchmarks, and tok/s reports end-to-end throughput at batch size 16 on a single H100 GPU. Within each model and regime, \textbf{bold} marks the two highest values and shading marks the best. Dense denotes the original model with no experts skipped.}
\label{tab:main_all}
\end{table}

\subsection{Experimental Setup}
\label{sec:experimental_setup}

\textbf{Models.} We evaluate BASE on three instruction-tuned MoE models with different routing
configurations. Qwen3-30B-A3B-Instruct-2507~\citep{yang2025qwen3technicalreport} contains 48
MoE layers, 128 routed experts per layer, and activates 8 routed experts per
token. Qwen1.5-MoE-A2.7B-Chat~\citep{bai2023qwen} contains 24 MoE layers, 60
routed experts, and activates 4 routed experts per token.
DeepSeek-V2-Lite-Chat~\citep{deepseekai2024deepseekv2strongeconomicalefficient}
contains 26 MoE layers, 64 routed experts, and activates 6 routed experts per
token.

\textbf{Benchmarks and metrics.} We evaluate all methods on the full, unsubsampled sets from eight generative
benchmarks: CMMLU~\citep{li2024cmmlumeasuringmassivemultitask}, BBH~\citep{suzgun2022challengingbigbenchtaskschainofthought},
MATH~\citep{hendrycks2021measuringmathematicalproblemsolving}, BoolQ~\citep{clark2019boolqexploringsurprisingdifficulty},
GSM8K~\citep{cobbe2021trainingverifierssolvemath}, MBPP~\citep{austin2021programsynthesislargelanguage},
MATH-401~\citep{yuan2023large}, and HumanEval~\citep{chen2021evaluatinglargelanguagemodels}. We
report accuracy for CMMLU, BBH, MATH, BoolQ, GSM8K, and MATH-401, pass rate for
MBPP, and pass@1 for HumanEval. The overall score is the average of the
eight benchmark scores.

\textbf{Baselines and implementation.} We compare BASE with four batch-aware expert-selection methods,
SERE~\citep{wu2026sere}, OEA~\citep{oncescu2025opportunistic},
Lynx~\citep{gupta2024lynx}, and ExFold~\citep{wu2026exfold}. All methods are
served with vLLM 0.9.2 (V0 engine) in bf16 with batch size 16 and
identical prompting and sampling settings: temperature 0.7, top-$p$ 0.8,
top-$k$ 20, and seed 0. Following common
practice~\citep{oncescu2025opportunistic,gupta2024lynx}, all
methods run prefill with the dense model and apply expert selection only during
decoding, where inference is memory-bound and cost grows with the number of
experts fetched. We report variation across seeds in
Appendix~\ref{app:seed_variance}. Each
run uses a single NVIDIA H100 80GB GPU. During offline calibration, BASE fits
its linear expert-energy predictors on a subset of the FineWeb-Edu sample-10BT
training split~\citep{penedo2024finewebdatasetsdecantingweb}. The selected text
is packed into 2,048-token sequences, totaling approximately 2.5 million
calibration tokens per model. The ridge regressions are solved in closed form. On one
NVIDIA RTX PRO 6000 Blackwell Max-Q GPU, calibration takes 38.9 minutes for
Qwen3-30B-A3B, 17.4 minutes for DeepSeek-V2-Lite, and 9.7 minutes for
Qwen1.5-MoE-A2.7B. The predictors can be calibrated independently and in
parallel across layers and experts, and our implementation processes them in
groups of eight layers. Further details are provided in
Appendix~\ref{app:calibration_details}.
BASE uses the backfill procedure described in
Section~\ref{sec:inference_procedure}. We ablate this routing rule and the expert-selection rule in
Sections~\ref{sec:fill_rule_ablation}
and~\ref{sec:selection_score_ablation}, respectively.

\textbf{Comparison protocol.} Because the methods use different expert-selection strategies, we compare them
at matched decoding speed and measure model quality. SERE's $S$
and OEA's $k_0$ both specify how many top-ranked experts are retained per token
before their union forms the initial batch active set; the methods differ in
how they fill the remaining routing slots. SERE also has a similarity threshold
($\rho$ in \citealp{wu2026sere}) that fetches additional experts when no active
expert is similar enough, improving quality but lowering decoding speed; to
keep budgets matched, we set $\rho=0$ for all SERE runs. We define two
operating regimes.
The \emph{restricted} regime uses SERE $S=1$ and OEA $k_0=1$ as reference
settings, while the \emph{relaxed} regime uses $S=2$ and $k_0=2$. For each
model, we select the BASE budget and the Lynx and ExFold settings that place
them in the corresponding decoding-speed range. Lynx's $\alpha,\beta$ and ExFold's
$D,P$ are method-specific hyperparameters that control the accuracy--speed
trade-off; we set them to match the decoding speed of the union-based SERE and
OEA settings in each regime. These settings are fixed
before quality is evaluated.

Throughput is measured end to end, including prefill, on the same 16 prompts
and the same machine using a single NVIDIA H100 80GB GPU. Every sequence is
forced to generate exactly 384 tokens with end-of-sequence termination
disabled, preventing any method from gaining throughput by producing shorter
outputs. We report the best of three runs after warm-up.

\subsection{Main Results}
\label{sec:main_results}

Table~\ref{tab:main_all} reports the main quality and throughput results. Under
the restricted regime, BASE achieves the highest average quality and
throughput on all three architectures. On Qwen3-30B-A3B, BASE reaches an average score
of $76.38$, compared with $46.87$ for the strongest baseline. The corresponding
gains are $2.96$ points on Qwen1.5-MoE-A2.7B and $6.61$ points on DeepSeek-V2-Lite,
showing that predicted removal error is especially useful when only a small
expert set can be fetched. As more experts become available, the quality
differences narrow: BASE exceeds the strongest baseline by $0.75$, $0.78$, and
$0.58$ points on Qwen3-30B-A3B, Qwen1.5-MoE-A2.7B, and DeepSeek-V2-Lite, respectively. BASE also
maintains the highest throughput in the relaxed regime. Overall, BASE provides
the best average quality and throughput across both operating regimes and all
three evaluated architectures.

\begin{table}[!t]
\centering
\begin{minipage}[t]{0.48\linewidth}
    \vspace{0pt}
    \centering
    \footnotesize
    \setlength{\tabcolsep}{1.5pt}
    \renewcommand{\arraystretch}{0.9}
    \begin{tabular}{lrr}
        \toprule
        Selection score & \makecell{Retained\\energy (\%)} & \makecell{Mean\\accuracy} \\
        \midrule
        Top-1 union & 85.91  & 58.68 \\
        Router weight sum         & 82.61  & 42.87 \\
        Square weight sum         & 90.98  & 55.50 \\
        Static expert energy      & 92.65  & 64.46 \\
        BASE linear predictor     & 96.47  & 73.50 \\
        Nonlinear predictor       & 98.20  & 72.07 \\
        Oracle energy             & 100.00 & 73.58 \\
        \bottomrule
    \end{tabular}
    \captionsetup{skip=4pt}
    \captionof{table}{Selection-rule ablation on Qwen3-30B-A3B with active-set size matched to the top-1 union. Retained energy measures the oracle energy preserved by the selected set.}
    \label{tab:selection_score_ablation}
\end{minipage}%
\hspace{0.02\linewidth}%
\begin{minipage}[t]{0.50\linewidth}
    \vspace{0pt}
    \centering
    \footnotesize
    \setlength{\tabcolsep}{2pt}
    \renewcommand{\arraystretch}{1.25}
    \begin{tabular}{@{}l@{\hspace{1.5pt}}r@{\hspace{2pt}}r@{\hspace{2pt}}r@{\hspace{2pt}}r@{\hspace{2pt}}r@{\hspace{2pt}}r@{}}
        \toprule
        Fill rule & HEval & GSM8K & MATH & BBH & Avg. & tok/s \\
        \midrule
        Drop         & 68.90 & 86.00 & 72.00 & 69.91 & 74.20 & 1473.4 \\
        Backfill     & \textbf{72.56} & \textbf{88.00} & \textbf{73.00} & \textbf{72.22} & \textbf{76.45} & \textbf{1479.3} \\
        Substitution & 7.93 & 36.00 & 22.50 & 43.52 & 27.49 & 1477.0 \\
        \bottomrule
    \end{tabular}
    \captionsetup{skip=4pt}
    \captionof{table}{Fill-rule ablation on Qwen3-30B-A3B using the BASE linear selector with $M=10$. Drop removes unavailable contributions, backfill uses each token's next-preferred experts in the active set, and substitution redirects each unavailable expert to its most similar active expert. Avg. is the average across the four benchmarks.}
    \label{tab:fill_rule_ablation}
\end{minipage}
\end{table}

\subsection{Ablation: Selection Rule}
\label{sec:selection_score_ablation}

We evaluate selection criteria on Qwen3-30B-A3B with $B=16$, dense prefill, and
the same backfill procedure and evaluation examples for every method. We report
mean accuracy across seven benchmarks using our Hugging Face implementation
on fixed subsets: all 164 HumanEval problems, the first 200
problems of MATH, GSM8K, BoolQ, MBPP, and MATH-401, and the first 8 problems of
each of the 27 BBH subtasks. We match expert-loading cost at every decode position by setting the budget to
the size of the tokens' top-1 union and requiring every method to select the
same number of experts. Thus, the
methods differ only in how experts are ranked.

We compare the top-1 union, the sum of router weights, and the sum of squared
router weights against scores that also use expert output energy. Static expert
energy replaces the token-specific energy in our removal score with each
expert's mean calibration energy, while BASE uses the prediction from
Section~\ref{sec:energy_prediction}. We also include a nonlinear
predictor as a higher-capacity variant. We additionally use the true expert output energies to form an oracle ranking
under the same budget. This oracle is not deployable because it requires
executing all routed experts before selection. Retained energy measures the
fraction of oracle contribution energy contained in the selected set, normalized
by the energy retained by the oracle set at the same decode position.
Appendix~\ref{app:selection_rule_ablation} gives the full experimental setup and
the formal definitions of every score, the oracle, and retained energy.

Table~\ref{tab:selection_score_ablation} shows that router scores alone do not
reliably identify the experts whose removal most affects quality. BASE increases
retained energy from $90.98\%$ for squared router weights to $96.47\%$, while
mean accuracy rises from $55.50\%$ to $73.50\%$, close to the oracle value of
$73.58\%$. The nonlinear predictor retains slightly more energy but does not
improve accuracy, so we use the linear predictor.

\subsection{Ablation: Fill Rule}
\label{sec:fill_rule_ablation}

Table~\ref{tab:fill_rule_ablation} isolates how unavailable experts should be
handled after selection. Using Qwen3-30B-A3B with $M=10$, we hold the BASE linear
selector and active set fixed and compare three fill rules, using the same
implementation on all 164 HumanEval problems, the first 200 GSM8K and MATH
problems, and the first 8 problems of each of the 27 BBH subtasks.
Dropping simply
removes unavailable contributions. Following SERE~\citep{wu2026sere},
substitution redirects each unavailable expert to the most similar expert in
the active set according to SERE's calibrated expert-similarity metric, while
preserving the original routing weight. Substitution reduces average accuracy
to $27.49$, consistent with Figure~\ref{fig:expert_geometry}(a), where
co-routed expert outputs are nearly orthogonal and contain little shared
directional information that would allow one expert's output to replace
another. Backfill instead uses each token's next-preferred experts within the
active set and improves average accuracy from $74.20$ to $76.45$. Although
backfill performs additional computation, these experts have already been
loaded and thus introduce no additional expert-weight transfers. The added
FLOPs therefore have negligible effect on throughput in the memory-bound
decoding regime.

\subsection{Ablation: Effect of Decode Batch Size}
\label{sec:batch_size_ablation}

\begin{figure}[t]
\centering
\includegraphics[width=\linewidth]{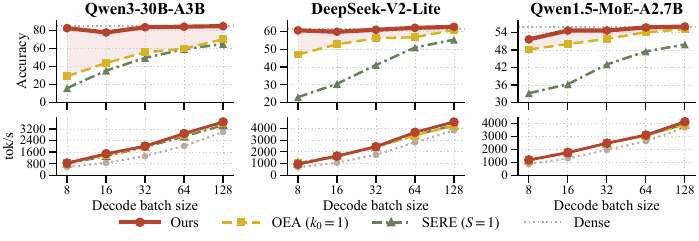}
\caption{Effect of decode batch size on quality and throughput across three MoE
architectures. The top row reports the average across HumanEval, MBPP,
MATH-401, and BoolQ; the bottom row reports token throughput. The dotted gray
line denotes the dense model. We vary $B$ from 8 to 128, and BASE's expert
budget is speed-matched to the faster baseline at each batch size.
Appendix~\ref{app:batch_size_sweep} gives the budgets and further details.}
\label{fig:bsweep}
\end{figure}

Figure~\ref{fig:bsweep} evaluates whether BASE's advantage depends on the
decode batch size. We sweep $B\in\{8,16,32,64,128\}$ across all three
architectures and compare against the restricted SERE $S=1$ and OEA $k_0=1$
settings. At each batch size, we first measure both baselines on the same H100
and use the faster token throughput as the speed target. We then choose the
BASE expert budget to match this target, fix that budget, and evaluate
quality. Thus, the comparison controls for serving
speed rather than expert count.
At matched throughput, BASE maintains relatively stable quality across the
full range of batch sizes, while SERE and OEA degrade more sharply as the batch
becomes smaller. BASE's quality advantage is therefore largest at smaller
batch sizes and remains positive as the batch grows. These results show that
BASE's quality advantage at comparable serving speed is consistent across
batch sizes and is not specific to the $B=16$ setting used in the main
evaluation.

\section{Conclusion}
\label{sec:conclusion}

MoE sparsity at the token level does not guarantee efficient batched decoding,
since concurrent requests can collectively activate many experts. We
introduced BASE, which selects experts under a fixed batch-level budget
according to their predicted effect on the MoE-layer output. Expert selection
remains robust after discarding cross-expert terms, allowing BASE to rank
experts independently using removal costs estimated by a lightweight linear
predictor. At tight expert budgets, BASE improves average accuracy by
$3.0$--$29.5$ points over existing baselines at comparable speed. At relaxed
budgets, it achieves comparable quality while maintaining higher throughput.
Relative to dense inference, relaxed BASE settings are $22$--$60\%$ faster
while remaining within $0.4$ accuracy points of dense quality; the most
constrained setting is $73\%$ faster with a $6.1$-point loss. One potential
limitation is that BASE requires more offline calibration time and compute
than simpler baselines.

\section*{Acknowledgements}
This work was supported in part by the National Center on Generative AI for
Uplifting STEM+C Education (GENIUS Center) and by the United States Air Force
under Contract No.~FA8750-23-C-0518. The authors also gratefully acknowledge
the computing resources provided by the NVIDIA Academic Grant Program and
Lambda Cloud.

\bibliography{references}
\bibliographystyle{iclr2027_conference}

\clearpage
\appendix
\section{Appendix}
\label{app:additional_results}

\subsection{Calibration Details}
\label{app:calibration_details}
\textbf{Data.} We calibrate BASE using educational web text from
\texttt{HuggingFaceFW/fineweb-edu}, configuration \texttt{sample-10BT}, training
split. We
read the data in deterministic stream order and retain documents containing
more than 128 tokens under the Qwen1.5 tokenizer. This produces 2,708 documents.
We reserve the final fifth of the documents for validation before packing, so
the training and validation splits are disjoint by document. Using the Qwen
tokenizer, the training split contains 1,199 sequences and 2,455,552 tokens,
with 239 additional validation sequences. Using the DeepSeek tokenizer, the
same text produces 1,230 training sequences and 2,519,040 tokens, with 246
validation sequences. All sequences contain 2,048 tokens. The validation
sequences are used only for the retained-energy analysis and never for fitting
the predictors.

\textbf{Streaming sufficient statistics.} For every MoE layer and routed
expert, the calibration target is
$y=\log\|E_u(z)\|_2^2$, and the input is the normalized hidden state
$\bar z=z/\|z\|_2$. We process batches of eight sequences using teacher-forced
forward passes. For each routed token-expert pair, we run the expert and compute
its true output energy. We then update the sufficient statistics used for ridge
regression: the sample count, first moments, Gram matrix, and input-target
cross-moment. Token-level hidden states and expert outputs are discarded after
every batch, so calibration does not create an activation cache. Gram matrices
are accumulated in fp32, while vector and scalar statistics use fp64.

The predictors can be calibrated independently and in parallel across layers
and experts. Each predictor depends only on the sufficient statistics collected
for its own layer-expert pair, not on coefficients fitted for any other layer
or expert. Our implementation processes groups of eight layers.

\textbf{Closed-form ridge fitting.} For each expert, we center the accumulated
statistics and solve
\begin{equation}
    a_u(\lambda)
    =
    \left(G_{c,u}+\lambda I\right)^{-1}q_{c,u},
    \qquad
    b_u(\lambda)
    =
    \bar y_u-a_u(\lambda)^\top\bar z_u,
\end{equation}
where $G_{c,u}$ and $q_{c,u}$ are the centered Gram matrix and cross-moment.
One eigendecomposition per expert evaluates 25 values of $\lambda$, logarithmically
spaced from $10^{-4}$ to $10^2$. We select $\lambda$ separately for each expert
by marginal likelihood. Exponentiating the predicted log-energy can
underestimate the mean output energy, so we apply an expert-specific correction
for this bias. The correction is folded into the deployed intercept, so the
inference-time predictor retains the form in
Equation~\ref{eq:log_energy_predictor}.

\begin{center}
\centering
\small
\setlength{\tabcolsep}{7pt}
\begin{tabular}{lrrr}
\toprule
\textbf{Model} & \textbf{Training tokens} & \textbf{Data passes} & \textbf{Time} \\
\midrule
Qwen3-30B-A3B       & 2,455,552 & 6 & 38.9 min \\
DeepSeek-V2-Lite    & 2,519,040 & 4 & 17.4 min \\
Qwen1.5-MoE-A2.7B  & 2,455,552 & 3 &  9.7 min \\
\bottomrule
\end{tabular}
\captionof{table}{Offline calibration cost on one NVIDIA RTX PRO 6000
Blackwell Max-Q GPU with 96 GB of memory.}
\label{tab:calibration_cost}
\end{center}

The reported times begin with the first statistics-collection pass and end
when the fitted coefficients are written. They include all forward passes,
expert recomputation, and ridge solves, but exclude model loading,
tokenization, and final artifact packaging. Calibration is performed once per
model before deployment; no recalibration occurs during serving.

\subsection{Cross-Term Bias: Formulation and Fitting}
\label{app:crossterm_fit}

This section describes the one-parameter fits shown by the dashed curves in
Figure~\ref{fig:expert_geometry}(b). We compute the exact-to-additive error
ratio $\rho_t$ using the ground-truth outputs of all experts routed to each
token. Therefore, this analysis contains no error from the expert-energy
predictor.

\begin{center}
\centering
\includegraphics[width=\textwidth]{figures/fig1.pdf}
\captionof*{figure}{Figure~\ref{fig:expert_geometry} reproduced here for
convenience.}
\end{center}

We use the same three model checkpoints as in the main experiments and
evaluate them on 16 held-out FineWeb-Edu sequences, each containing 2,048
tokens. At each sequence position, the corresponding 16 tokens form one
decode batch. The active set is constructed as the union of the top-1 expert
selected by each token in the batch. Shared experts are always active and are
excluded from this analysis.

Recall that the exact-to-additive error ratio is

\begin{equation}
\rho_t
=
\frac{
\left\|\sum_{u\in U_t}a_{t,u}\right\|_2^2
}{
\sum_{u\in U_t}\|a_{t,u}\|_2^2
},
\end{equation}

where $U_t=R_t\setminus A_\ell$ is the set of experts originally routed to
token $t$ but omitted from the active set.

For an omitted expert $u$, let
$m_{t,u}=w_{t,u}\lVert E_u(z_t)\rVert_2$ denote the magnitude of its weighted
contribution. Expanding the exact reconstruction error gives

\begin{equation}
\left\|\sum_{u\in U_t}a_{t,u}\right\|_2^2
=
\sum_{u\in U_t}m_{t,u}^2
+
\sum_{\substack{u,v\in U_t\\u\neq v}}
m_{t,u}m_{t,v}c_{t,uv},
\label{eq:appendix_crossterm_expansion}
\end{equation}

where $c_{t,uv}$ is the cosine similarity between two omitted expert
contributions. We average these similarities across omitted pairs, weighting
each pair by the product of its contribution magnitudes:

\begin{equation}
\bar c_t
=
\frac{
\sum_{u\neq v}m_{t,u}m_{t,v}c_{t,uv}
}{
\sum_{u\neq v}m_{t,u}m_{t,v}
},
\label{eq:weighted_mean_cosine}
\end{equation}

and define their effective count as

\begin{equation}
|U_t|_{\mathrm{eff}}
=
\frac{
\left(\sum_{u\in U_t}m_{t,u}\right)^2
}{
\sum_{u\in U_t}m_{t,u}^2
}.
\label{eq:effective_omitted_count}
\end{equation}

These definitions give the exact identity

\begin{equation}
\rho_t-1
=
\bar c_t\left(|U_t|_{\mathrm{eff}}-1\right).
\label{eq:crossterm_exact_identity}
\end{equation}

Thus, the deviation from additivity depends on both the alignment of the
omitted contributions and the number of omitted contributions with meaningful
magnitude. The effective count satisfies
$1\leq |U_t|_{\mathrm{eff}}\leq |U_t|$. It equals $|U_t|$ when all omitted
contributions have the same magnitude and approaches one when a single
contribution dominates. When only one expert is omitted,
$|U_t|_{\mathrm{eff}}=1$, so $\rho_t=1$ exactly.

To obtain the dashed curves in Figure~\ref{fig:expert_geometry}(b), we replace
the token-dependent cosine $\bar c_t$ with one fitted constant $\bar c$ for
each model:

\begin{equation}
\rho_t
\approx
1+\bar c\left(|U_t|_{\mathrm{eff}}-1\right).
\label{eq:crossterm_one_parameter_fit}
\end{equation}

We group samples according to the number of omitted experts,
$k=|U_t|$, and retain bins with at least 100 samples. For each bin, we compute
the mean exact ratio $\bar\rho_k$ and mean effective count $\bar e_k$. We then
fit $\bar c$ using equally weighted least squares with no intercept:

\begin{equation}
\bar c
=
\frac{
\sum_k(\bar\rho_k-1)(\bar e_k-1)
}{
\sum_k(\bar e_k-1)^2
}.
\label{eq:crossterm_fit_solution}
\end{equation}

The horizontal axis of Figure~\ref{fig:expert_geometry}(b) remains the literal
count $k=|U_t|$. The effective count enters only the dashed curve: for the
tokens in bin $k$, we average $|U_t|_{\mathrm{eff}}$ to obtain $\bar e_k$ and
plot $1+\bar c(\bar e_k-1)$ at $x=k$. The two counts differ because omitted
contributions can have unequal magnitudes. For example, Qwen3-30B-A3B tokens with
$|U_t|=7$ have a mean effective count of only $5.48$, causing the fitted curve
to bend below a line based directly on $|U_t|$.

\begin{center}
\centering
\small
\setlength{\tabcolsep}{6pt}
\renewcommand{\arraystretch}{1.1}
\begin{tabular}{lr}
\toprule[1.4pt]
Model & Fitted $\bar c$ \\
\midrule[0.75pt]
Qwen1.5-MoE-A2.7B & 0.021 \\
DeepSeek-V2-Lite  & 0.026 \\
Qwen3-30B-A3B     & 0.032 \\
\bottomrule[1.4pt]
\end{tabular}
\captionof{table}{Fitted constants for the one-parameter cross-term model in
Figure~\ref{fig:expert_geometry}(b).}
\label{tab:crossterm_fit}
\end{center}

The fitted curves remain within $0.65\%$ of the observed mean $\rho_t$ for
every plotted bin. This confirms that cross terms introduce a small and
predictable count-dependent bias in the magnitude of the additive error.

\subsection{Robustness Across Sampling Seeds}
\label{app:seed_variance}

We repeat the headline Qwen3-30B-A3B comparisons using seeds $0$, $1$, and
$2$. We evaluate the restricted BASE configuration with $M=10$ against OEA
with $k_0=1$ and SERE with $S=1$. We also evaluate the relaxed BASE
configuration with $M=16$ against OEA with $k_0=2$.

Each seed controls both the vLLM engine and generation sampling, while all
other inference settings remain unchanged. We use the same model, calibration
data, prompts, sampling parameters, and H100 hardware as in the main
evaluation, with a decode batch size of $16$. This ablation covers five
benchmarks: HumanEval, MBPP, MATH-401, GSM8K, and MATH-200. HumanEval, MBPP,
MATH-401, and GSM8K use their full evaluation sets. To limit computational
cost, MATH-200 uses the first 200 examples from the MATH test set. For every
benchmark with an existing main-table reference, the seed-$0$ run reproduces
the corresponding outputs item for item, confirming that this study uses the
same inference implementation.

\begin{center}
\centering
\footnotesize
\setlength{\tabcolsep}{2.8pt}
\renewcommand{\arraystretch}{1.18}
\resizebox{\textwidth}{!}{%
\begin{tabular}{clccccc}
\toprule[1.4pt]
& \textbf{Methods \textbackslash{} Tasks}
& \textbf{HumanEval} & \textbf{MATH-200} & \textbf{MBPP}
& \textbf{MATH-401} & \textbf{GSM8K} \\
\midrule[0.75pt]
\multirow{3}{*}{\rotatebox[origin=c]{90}{\makecell[c]{\scriptsize\textbf{Restricted}\\[-1pt]\scriptsize$\approx S\!=\!1$}}}
& BASE (ours) \textcolor{gray}{$_{M=10}$}
& \cellcolor{besttint}\makecell{\bfseries $73.58 \pm 1.54$\\[-1pt]{\fontsize{4.5}{5}\selectfont\textcolor{gray}{$(75.00,\,71.95,\,73.78)$}}}
& \cellcolor{besttint}\makecell{\bfseries $68.50 \pm 3.28$\\[-1pt]{\fontsize{4.5}{5}\selectfont\textcolor{gray}{$(71.50,\,69.00,\,65.00)$}}}
& \cellcolor{besttint}\makecell{\bfseries $60.27 \pm 0.81$\\[-1pt]{\fontsize{4.5}{5}\selectfont\textcolor{gray}{$(59.80,\,61.20,\,59.80)$}}}
& \cellcolor{besttint}\makecell{\bfseries $81.21 \pm 1.15$\\[-1pt]{\fontsize{4.5}{5}\selectfont\textcolor{gray}{$(80.55,\,82.54,\,80.55)$}}}
& \cellcolor{besttint}\makecell{\bfseries $88.78 \pm 0.40$\\[-1pt]{\fontsize{4.5}{5}\selectfont\textcolor{gray}{$(89.01,\,89.01,\,88.32)$}}} \\
& OEA \textcolor{gray}{$_{k_0=1}$}
& \makecell{$43.49 \pm 1.27$\\[-1pt]{\fontsize{4.5}{5}\selectfont\textcolor{gray}{$(43.90,\,44.51,\,42.07)$}}}
& \makecell{$24.67 \pm 2.31$\\[-1pt]{\fontsize{4.5}{5}\selectfont\textcolor{gray}{$(26.00,\,22.00,\,26.00)$}}}
& \makecell{$10.13 \pm 1.10$\\[-1pt]{\fontsize{4.5}{5}\selectfont\textcolor{gray}{$(9.60,\,9.40,\,11.40)$}}}
& \makecell{$36.24 \pm 1.51$\\[-1pt]{\fontsize{4.5}{5}\selectfont\textcolor{gray}{$(34.66,\,36.41,\,37.66)$}}}
& \makecell{$32.80 \pm 0.43$\\[-1pt]{\fontsize{4.5}{5}\selectfont\textcolor{gray}{$(33.28,\,32.45,\,32.68)$}}} \\
& SERE \textcolor{gray}{$_{S=1}$}
& \makecell{$13.01 \pm 3.35$\\[-1pt]{\fontsize{4.5}{5}\selectfont\textcolor{gray}{$(16.46,\,12.80,\,9.76)$}}}
& \makecell{$19.33 \pm 1.76$\\[-1pt]{\fontsize{4.5}{5}\selectfont\textcolor{gray}{$(21.00,\,17.50,\,19.50)$}}}
& \makecell{$8.73 \pm 0.70$\\[-1pt]{\fontsize{4.5}{5}\selectfont\textcolor{gray}{$(9.40,\,8.80,\,8.00)$}}}
& \makecell{$39.40 \pm 1.50$\\[-1pt]{\fontsize{4.5}{5}\selectfont\textcolor{gray}{$(40.90,\,37.91,\,39.40)$}}}
& \makecell{$40.46 \pm 1.48$\\[-1pt]{\fontsize{4.5}{5}\selectfont\textcolor{gray}{$(38.97,\,41.93,\,40.49)$}}} \\
\cmidrule[0.6pt](lr){1-7}
\multirow{2}{*}{\rotatebox[origin=c]{90}{\makecell[c]{\scriptsize\textbf{Relaxed}\\[-1pt]\scriptsize$\approx S\!=\!2$}}}
& BASE (ours) \textcolor{gray}{$_{M=16}$}
& \cellcolor{besttint}\makecell{\bfseries $87.81 \pm 1.61$\\[-1pt]{\fontsize{4.5}{5}\selectfont\textcolor{gray}{$(87.20,\,86.59,\,89.63)$}}}
& \cellcolor{besttint}\makecell{\bfseries $79.17 \pm 1.61$\\[-1pt]{\fontsize{4.5}{5}\selectfont\textcolor{gray}{$(81.00,\,78.00,\,78.50)$}}}
& \cellcolor{besttint}\makecell{\bfseries $77.07 \pm 0.31$\\[-1pt]{\fontsize{4.5}{5}\selectfont\textcolor{gray}{$(77.00,\,77.40,\,76.80)$}}}
& \cellcolor{besttint}\makecell{\bfseries $81.80 \pm 0.66$\\[-1pt]{\fontsize{4.5}{5}\selectfont\textcolor{gray}{$(81.55,\,81.30,\,82.54)$}}}
& \cellcolor{besttint}\makecell{\bfseries $89.46 \pm 0.59$\\[-1pt]{\fontsize{4.5}{5}\selectfont\textcolor{gray}{$(90.14,\,89.08,\,89.16)$}}} \\
& OEA \textcolor{gray}{$_{k_0=2}$}
& \makecell{$83.34 \pm 0.70$\\[-1pt]{\fontsize{4.5}{5}\selectfont\textcolor{gray}{$(82.93,\,82.93,\,84.15)$}}}
& \makecell{$68.00 \pm 3.12$\\[-1pt]{\fontsize{4.5}{5}\selectfont\textcolor{gray}{$(69.00,\,70.50,\,64.50)$}}}
& \makecell{$27.67 \pm 1.45$\\[-1pt]{\fontsize{4.5}{5}\selectfont\textcolor{gray}{$(28.60,\,28.40,\,26.00)$}}}
& \makecell{$73.07 \pm 1.09$\\[-1pt]{\fontsize{4.5}{5}\selectfont\textcolor{gray}{$(71.82,\,73.57,\,73.82)$}}}
& \makecell{$76.70 \pm 0.36$\\[-1pt]{\fontsize{4.5}{5}\selectfont\textcolor{gray}{$(77.10,\,76.57,\,76.42)$}}} \\
\bottomrule[1.4pt]
\end{tabular}%
}
\captionof{table}{Robustness to sampling seeds on Qwen3-30B-A3B. Each entry reports
the mean $\pm$ sample standard deviation over three sampling seeds, followed in
parentheses by the individual results for seeds $0$, $1$, and $2$. BASE uses
the ridge-only linear predictor at the restricted budget $M=10$ and relaxed
budget $M=16$. Bold and shading indicate the best mean within each operating
regime.}
\label{tab:seed_variance_qwen3}
\end{center}

Table~\ref{tab:seed_variance_qwen3} reports the mean and sample standard
deviation across the three seeds, followed by the individual seed results in
parentheses. BASE remains consistently stronger than its matched-speed
baselines. Its sample standard deviation ranges from $0.31$ to $1.61$ on the
four full benchmarks and from $1.61$ to $3.28$ on MATH-200. The closest
comparison occurs against OEA on HumanEval in the relaxed regime, where BASE
retains a $4.47$-point advantage in mean accuracy.

\subsection{Decode Latency Versus Expert Budget}
\label{app:latency_budget}

We measure how the fixed expert budget affects decode latency on all three
architectures. Measurements use vLLM 0.9.2 with the V0 engine, bf16 precision,
decode batch size $16$, and one NVIDIA RTX PRO 6000 Blackwell Max-Q GPU. We vary
$M\in\{4,8,12,16,20,24,32\}$. For each setting, we run three trials and report
the lowest decode time per output token (TPOT), the quantity directly modeled
by our linear latency analysis.

We measure end-to-end decode time without adding profiling operations to the
decode loop. Active-set size is measured separately to avoid affecting
latency. In this experiment, the selector admits exactly $M$ experts at each
layer and decode step. The realized active-set size counts how many of those
experts receive at least one token and are therefore dispatched. It can be
smaller than $M$ because the tokens in a batch may collectively request fewer
than $M$ distinct experts, leaving some admitted experts unused.

\begin{center}
\centering
\includegraphics[width=\textwidth]{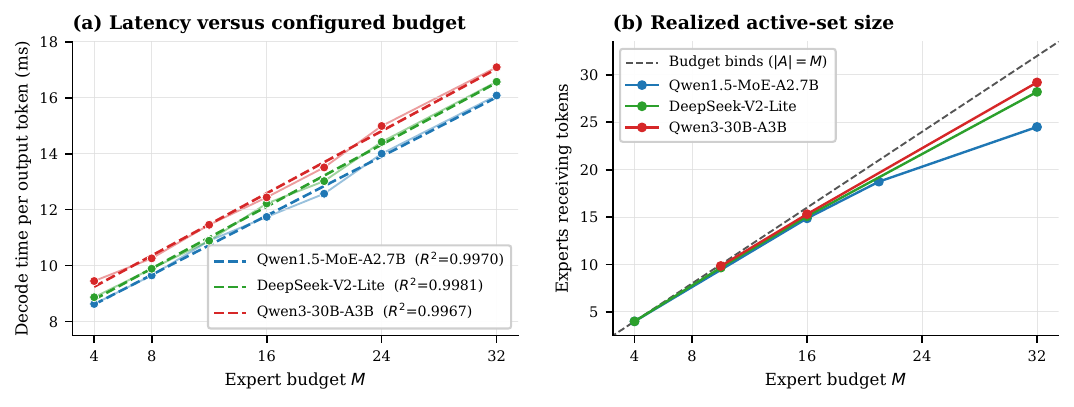}
\captionof{figure}{Decode latency as a function of the fixed expert budget.
\textbf{(a)} Decode time per output token as $M$ increases. Dashed lines show
least-squares fits over $4\leq M\leq32$.
\textbf{(b)} Number of experts that receive tokens under each configured
budget. The dashed reference denotes $|A|=M$. At larger budgets, the batch may
request fewer than $M$ distinct experts, particularly for Qwen1.5-MoE-A2.7B.}
\label{fig:latency_budget}
\end{center}

\begin{center}
\centering
\small
\setlength{\tabcolsep}{5pt}
\renewcommand{\arraystretch}{1.1}
\begin{tabular}{lrrrrrr}
\toprule[1.4pt]
Model & Experts & \makecell{MoE\\layers} & $R^2$ &
\makecell{Intercept\\(ms)} & \makecell{Slope\\($\mu$s/$M$)} &
\makecell{Slope/layer\\($\mu$s)} \\
\midrule[0.75pt]
Qwen3-30B-A3B       & 128 & 48 & 0.9967 & 8.13 & 278.1 & 5.79 \\
Qwen1.5-MoE-A2.7B  &  60 & 24 & 0.9970 & 7.55 & 264.5 & 11.02 \\
DeepSeek-V2-Lite   &  64 & 26 & 0.9981 & 7.68 & 276.8 & 10.65 \\
\bottomrule[1.4pt]
\end{tabular}
\captionof{table}{Linear fits between decode latency and expert budget. We fit
$t(M)=a+bM$ over $M\in\{4,8,12,16,20,24,32\}$. The slope reports the increase
in decode time per output token from adding one unit to the expert budget.}
\label{tab:latency_budget_regression}
\end{center}

Figure~\ref{fig:latency_budget}(a) shows a near-linear relationship between
decode latency and $M$, with $R^2$ between $0.9967$ and $0.9981$ across the
three architectures. This agrees with the finding of
\citet{oncescu2025opportunistic} that decode latency scales approximately
linearly with the number of active experts in the memory-bound regime. The
fitted slopes are also similar, ranging from $264.5$ to $278.1\,\mu$s per unit
of budget. Panel (b) shows that the realized active-set size remains close to
$M$ at the budgets used in our main experiments. Across the operating budgets
of $M=10$ to $21$, realized expert activation is between $89\%$ and $98\%$ of
$M$. Thus, within our evaluated operating range, the configured budget closely
approximates the number of experts that are actually dispatched.

\subsection{Custom GPU Kernels}
\label{app:custom_kernels}

BASE runs after the model's standard top-$K$ router and before expert execution.
For each token $t$, the routing stage provides its hidden state $z_t$, routed
expert IDs $R_t$, and router weights $w_{t,u}$. We normalize $z_t$ once and
reuse it across the token's $K$ routed experts. For each routed expert $u$, the
scoring kernel combines this token-specific input with the expert-specific
linear parameters $a_u$ and $b_u$ to predict its output energy,
$\widehat e_u(z_t)=\exp(a_u^\top\bar z_t+b_u)$. It then multiplies this
prediction by $w_{t,u}^2$ and aggregates the resulting removal costs across all
tokens routed to the same expert, producing one batch-level score $\widehat
D_u$ per expert. A second kernel selects the $M$ highest-scoring experts to
form an active set shared by the batch. Finally, the backfill kernel processes
each token's router ranking and selects its $K$ highest-ranked experts within
this active set. We implement this data flow using four fused Triton launches
and two lightweight elementwise operations, instead of roughly 20 separate
PyTorch launches per layer.

To isolate the cost introduced by BASE, we begin with vLLM's fused top-$K$
router and cumulatively add scoring, active-set selection, and backfill. The
measurements use hidden states and router logits captured from real decode
steps and the final linear-predictor implementation. Table~\ref{tab:kernel_profile_h100}
reports the incremental latency on a single H100 GPU at batch size $16$. The
total BASE overhead is measured directly from the complete routing function;
small differences from the sum of the individual stages result from rounding
and interactions between consecutive kernels.

\begin{center}
\centering
\footnotesize
\setlength{\tabcolsep}{3.2pt}
\renewcommand{\arraystretch}{1.1}
\resizebox{\textwidth}{!}{%
\begin{tabular}{lrrrrrrrr}
\toprule[1.4pt]
\textbf{Model} & $M$ &
\makecell{Standard\\router} &
\makecell{BASE\\scoring} &
\makecell{Expert\\selection} &
\textbf{Backfill} &
\makecell{Total BASE\\overhead} &
\makecell{Per decode\\step} &
\makecell{Share of\\decode step} \\
\midrule[0.75pt]
Qwen3-30B-A3B          & 10 & 10.8 & $+3.6$ & $+4.4$ & $+6.7$ & \textbf{$+14.9$} & 0.72 ms & 7.1\% \\
DeepSeek-V2-Lite   & 12 &  6.5 & $+9.8$ & $+2.3$ & $+5.8$ & \textbf{$+17.9$} & 0.47 ms & 5.1\% \\
Qwen1.5-MoE-A2.7B        & 14 &  9.1 & $+3.7$ & $+2.2$ & $+5.3$ & \textbf{$+11.3$} & 0.27 ms & 3.1\% \\
\bottomrule[1.4pt]
\end{tabular}
}
\captionof{table}{Incremental latency of BASE-specific routing operations in
$\mu$s per MoE layer at batch size $16$. The standard-router column contains
only vLLM's fused top-$K$ routing cost. The next three columns report the cost
added by each BASE stage. Per-step latency includes all MoE layers in the
corresponding model.}
\label{tab:kernel_profile_h100}
\end{center}

At batch size $16$, BASE adds $11.3$--$17.9\,\mu$s per MoE layer, corresponding
to $0.27$--$0.72$ ms per decode step and $3.1$--$7.1\%$ of the measured decode
time. Across batch sizes from $8$ to $128$, the overhead remains between
$11$ and $21\,\mu$s per layer, or $1$--$8\%$ of the decode step. Its relative
share decreases as the batch grows because expert computation increases much
more quickly than the routing overhead.

We verify the fused implementation against the complete deployed routing
function for every profiled model and batch size. The kernels admit exactly
$M$ experts, and the stage-by-stage path produces identical expert IDs and
routing weights to the complete function. Thus, the measurements decompose the
deployed BASE routing path rather than a separate profiling implementation.

\subsection{Selection-Rule Ablation: Setup and Score Definitions}
\label{app:selection_rule_ablation}

This section gives the full setup and the formal definition of each score for
the selection-rule ablation in Section~\ref{sec:selection_score_ablation}.

We evaluate the selection criteria on Qwen3-30B-A3B with decode batch size
$B=16$, dense prefill, and the same backfill procedure and evaluation examples
for every method. This ablation was run on a single NVIDIA RTX PRO 6000
Blackwell Max-Q GPU. During offline calibration, we fit the expert-energy
predictors using educational web text from the FineWeb-Edu sample-10BT
training split~\citep{penedo2024finewebdatasetsdecantingweb}. Documents are selected deterministically, filtered to contain
more than 128 tokens, and packed into 2,048-token sequences. We report mean
accuracy across HumanEval, MATH, GSM8K, BIG-Bench Hard, BoolQ, MBPP, and
MATH-401.

We match expert-loading cost separately at every decode position. Specifically,
the budget is the number of distinct experts in the tokens' top-1 union, and
every method selects exactly that many experts. For $B=16$, this produces a
mean active-set size of $13.34$, with values ranging from $1$ to $16$. Thus,
only the criterion used to rank experts changes across methods. We compare two
families: scores derived only from router information, and scores that also
incorporate information about expert outputs.

\textbf{Router-only selection.} The first family uses only information produced by the router. The \emph{top-1
union} retains every expert ranked first by at least one token, assigning
binary importance according to union membership. The \emph{router-weight sum} ranks
experts by the sum of their routing weights across the batch,
$S_u^{\mathrm{weight}}=\sum_{t:u\in R_t}w_{t,u}$. The \emph{squared
router-weight sum}
instead uses $S_u^{\mathrm{sq}}=\sum_{t:u\in R_t}w_{t,u}^2$, matching the
quadratic dependence on routing weights in our individual-damage objective.
None of these criteria uses information about the outputs produced by the
experts.

\textbf{Router and expert-output selection.} The second family augments the squared router-weight sum with information about
expert output magnitude. The \emph{static-energy} score uses
$S_u^{\mathrm{static}}=\sum_{t:u\in R_t}w_{t,u}^2\mu_u$, where
$\mu_u=\mathbb{E}_z[\|E_u(z)\|_2^2]$ is estimated during calibration and
remains fixed during inference. BASE replaces this constant with the
token-dependent estimate produced by our linear predictor,
$S_u^{\mathrm{BASE}}=\sum_{t:u\in R_t}w_{t,u}^2
\exp(a_u^\top\bar z_t+b_u)$. The \emph{nonlinear predictor} is a more
expensive variant that adds a shared nonlinear residual to this predictor.

\textbf{Oracle-energy selection and evaluation.} To separate prediction error from the underlying selection criterion, we
compute an oracle score from the experts' realized outputs. For each token
$t$, the true contribution energy of expert $u$ is
$w_{t,u}^2\|E_u(z_t)\|_2^2$. Aggregating this quantity across the batch gives
the oracle score
\begin{equation}
    D_u^\star
    =
    \sum_{t:u\in R_t}
    w_{t,u}^2\|E_u(z_t)\|_2^2,
    \label{eq:oracle_expert_score}
\end{equation}
where $R_t$ is the set of experts selected by the router for token $t$. This
score cannot be computed before selection without executing every routed
expert, which would defeat the purpose of reducing expert-weight transfers.
The resulting \emph{oracle-energy} selection is therefore not a deployable
baseline. Instead, it provides a
diagnostic upper bound under the same expert budget. Each method first uses
its own importance criterion to rank the experts and select $A^{(m)}$. After
the selection is fixed, we discard the method-specific scores and evaluate the
selected experts using $D_u^\star$. Oracle selection sorts the experts directly
by $D_u^\star$ under the same budget, producing $A^\star$. We define the energy
retained by method $m$ as
\begin{equation}
    \operatorname{ER}(m)
    =
    \frac{
        \sum_{u\in A^{(m)}}D_u^\star
    }{
        \sum_{u\in A^\star}D_u^\star
    }.
    \label{eq:energy_retained}
\end{equation}
Because $A^\star$ contains the experts with the largest oracle scores,
$0\leq\operatorname{ER}(m)\leq1$. Values closer to $1$ indicate that the
method selects an expert set more similar in retained energy to the oracle
selection. We compute this ratio independently at each decode position and
report its mean across all positions.

\subsection{Batch-Size Sweep: Experimental Details}
\label{app:batch_size_sweep}

This section gives the experimental details of the batch-size sweep in
Section~\ref{sec:batch_size_ablation} and Figure~\ref{fig:bsweep}. We sweep
$B\in\{8,16,32,64,128\}$ on all three architectures and compare BASE with the
restricted SERE ($S=1$) and OEA ($k_0=1$) settings. All runs use vLLM 0.9.2
(V0 engine) on a single H100 GPU with the prompts, sampling settings, and seed
of the main evaluation, and BASE uses the FineWeb-Edu calibration of
Appendix~\ref{app:calibration_details}. Quality is the mean accuracy over the
full HumanEval (164 problems), MBPP (500), MATH-401 (401), and BoolQ (3,270)
sets, with generation limits of 512 tokens for the first three and 2,048
tokens for BoolQ.

We set the BASE expert budget at each $B$ to match the token throughput of the
faster baseline at that batch size. For $B=8$, $16$, $32$, $64$, and $128$, this gives
$M=10$, $10$, $24$, $32$, and $48$ on Qwen3-30B-A3B, $M=10$, $12$, $19$, $24$, and $32$ on
DeepSeek-V2-Lite, and $M=6$, $14$, $21$, $34$, and $34$ on Qwen1.5-MoE-A2.7B.

Figure~\ref{fig:bsweep} reports end-to-end throughput from the timing harness of the main results on a single H100: the
batch is fixed to $B$, end-of-sequence tokens are ignored so that every
sequence runs to its full length, CUDA graphs are enabled, and we report the
best of three runs. The match is approximate: the faster baseline exceeds BASE
at four points: OEA by $2.8\%$ for Qwen3-30B-A3B at $B=8$, SERE by $9.0\%$ and $0.6\%$ for
DeepSeek-V2-Lite at $B=8$ and $B=16$, and OEA by $1.5\%$ for Qwen1.5-MoE-A2.7B at
$B=16$. At these points, the accuracy of BASE is higher by $53.3$, $37.6$,
$29.4$, and $4.7$ points, respectively.

\end{document}